\documentclass[conference,compsoc]{IEEEtran}
\usepackage{amsmath,graphicx}
\usepackage{multirow}
\usepackage{placeins}
\usepackage[font=small,skip=0pt]{caption}

\ifCLASSOPTIONcompsoc
  \usepackage[nocompress]{cite}
\else
  \usepackage{cite}
\fi

\ifCLASSINFOpdf
\else
\fi

\begin{document}
	\nocite{*}

\title{Simplified Long Short-term Memory \\ Recurrent Neural Networks:  part I}


\author{\IEEEauthorblockN{Atra Akandeh and Fathi M. Salem}
\IEEEauthorblockA{Circuits, Systems, and Neural Networks (CSANN) Laboratory \\
Computer Science and Engineering $||$ Electrical and Computer Engineering \\
Michigan State University\\
East Lansing, Michigan 48864-1226\\
akandeha@msu.edu; salemf@msu.edu }
}

\maketitle

\begin{abstract}
We present five variants of the standard Long Short-term Memory (LSTM) recurrent neural networks by uniformly reducing blocks of adaptive parameters in the gating mechanisms. For simplicity, we refer to these models as LSTM1, LSTM2, LSTM3, LSTM4, and LSTM5, respectively. Such parameter-reduced variants enable speeding up data training computations and would be more suitable for implementations onto constrained embedded platforms. We comparatively evaluate and verify our five variant models on the classical MNIST dataset and demonstrate that these variant models are comparable to a standard implementation of the LSTM model while using less number of parameters. Moreover, we observe that in some cases the standard LSTM's accuracy performance will drop after a number of epochs when using the ReLU nonlinearity; in contrast, however, LSTM3, LSTM4 and LSTM5 will retain their performance. 
\end{abstract}
 

%
\IEEEpeerreviewmaketitle

\section{Introduction}
Gated Recurrent Neural Networks (RNNs) have shown great success in processing data sequences in application such as speech recognition, natural language processing, and language translation. Gated RNNs are more powerful extension of the so-called simple RNNs. A simple RNN model is usually expressed using following equations:

\begin{equation}
	\begin{split}
		& h_t = \sigma(W_{hx} x_t + W_{hh} h_{t-1} + b_h) \\
		& y_t = W_{hy} h_t + b_y
	\end{split}
\end{equation}

\noindent where $W_{hx}, W_{hh}, b_h, W_{hy}, b_y$ are adaptive set of weights and $\sigma$ is a nonlinear bounded function. In the LSTM model the usual activation function has been replaced with a more equivalent complicated activation function, i.e. the hidden units are changed in such a way that the back propagated gradients are better behaved and permitting sustained gradient descent without vanishing to zero or growing unbounded \cite{lstm}. The LSTM RNN uses memory cells containing three gates: (i) an input (denoted by $i_t$), (ii) an output (denoted by $o_t$ and (iii) a forget ( denoted by $f_t$) gates. These gates collectively control signaling. Specially, the standard LSTM is expressed mathematically as
\begin{equation}
	\begin{split}
		& i_t = \sigma_{in}(W_i x_t + U_i h_{t-1} + b_i) \\
		& f_t = \sigma_{in}(W_f x_t + U_f h_{t-1} + b_f) \\
		& o_t = \sigma_{in}(W_o x_t + U_o h_{t-1} + b_o) \\
		& \tilde{c_t} = \sigma(W_c x_t + U_c h_{t-1} + b_c) \\
		& c_t= f_t \odot c_{t-1} + i_t \odot \tilde{c_t}\\
		& h_t = o_t \odot \sigma(c_t) 
	\end{split}
\end{equation}

\noindent where $\sigma_{in}$ is called inner activation (logistic) function which is bounded between 0 and 1, and $\odot$ denotes point-wise multiplication. The output layer of the LSTM model may be chosen to be as a linear map, namely,
\begin{equation}
	y_t = W_{hy} h_t + b_y
\end{equation}
LSTMs can be viewed as composed of the cell network and its 3 gating networks. LSTMs are relatively slow due to the fact that they have four sets of "weights," of which three are involved in the gating mechanism. In this paper we describe and demonstrate the comparative performance of five simplified LSTM variants by removing select blocks of adaptive parameters from the gating mechanism, and demonstrate that these variants are competitive alternate to the original LSTM model while requiring less computational cost.

\section{New Variants of the LSTM model}
LSTM uses gating mechanism to control the signal flow. It possess three gating signals driven by 3 main components, namely, the external input signal, the previous state, and a bias. We have proposed five variants of the LSTM model, aiming at reducing the number of (adaptive) parameters in each gate, and thus reduce computational cost \cite{salem_reduced}. The first three models have been demonstrated previously in initial experiments in \cite{lu}. In this work, we detail and demonstrate the comparative performance of the expanded 5 variants using the classical benchmark MNIST dataset formatted in sequence mappings experiments. Moreover, for modularity and ease in implementation, we apply the same changes to all three gates uniformly.

\subsection{LSTM1}
In this first model variant, input signals and their corresponding weights, namely, the terms $W_i x_t, W_f x_t, W_o x_t$ have been removed from the equations in the three corresponding gating signals. The resulting result model becomes
\begin{equation}
	\begin{split}
		& i_t = \sigma_{in}(U_i h_{t-1} + b_i) \\
		& f_t = \sigma_{in}(U_f h_{t-1} + b_f) \\
		& o_t = \sigma_{in}(U_o h_{t-1} + b_o) \\
		& \tilde{c_t} = \sigma(W_c x_t + U_c h_{t-1} + b_c) \\
		& c_t= f_t \odot c_{t-1} + i_t \odot \tilde{c_t}\\
		& h_t = o_t \odot \sigma(c_t) 
	\end{split}
\end{equation}

\subsection{LSTM2}
In this second model variant, the gates have no bias and no input signals $W_i x_t, W_f x_t, W_o x_t$. Only the state is used in the gating signals. This produces
\begin{equation}
	\begin{split}
		& i_t = \sigma_{in}(U_i h_{t-1}) \\
		& f_t = \sigma_{in}(U_f h_{t-1}) \\
		& o_t = \sigma_{in}(U_o h_{t-1}) \\
		& \tilde{c_t} = \sigma(W_c x_t + U_c h_{t-1} + b_c) \\
		& c_t= f_t \odot c_{t-1} + i_t \odot \tilde{c_t}\\
		& h_t = o_t \odot \sigma(c_t) 
	\end{split}
\end{equation}

\subsection{LSTM3}
In the third model variant, the only term in the gating signal is the (adaptive) bias. This model uses the least number of parameter among other variants.
\begin{equation}
	\begin{split}
		& i_t = \sigma_{in}(b_i) \\
		& f_t = \sigma_{in}(b_f) \\
		& o_t = \sigma_{in}(b_o) \\
		& \tilde{c_t} = \sigma(W_c x_t + U_c h_{t-1} + b_c) \\
		& c_t= f_t \odot c_{t-1} + i_t \odot \tilde{c_t}\\
		& h_t = o_t \odot \sigma(c_t) 
	\end{split}
\end{equation}

\subsection{LSTM4}
In the fourth model variant, the $U_i, U_f, U_o$ matrices have been replaced with the corresponding  $u_i, u_f, u_o$ vectors in LSTM2. The intent is to render the state signal with a point-wise multiplication. Thus, one reduces parameters while retain state feedback in the gatings.
\begin{equation}
	\begin{split}
		& i_t = \sigma_{in}(u_i  \odot h_{t-1}) \\
		& f_t = \sigma_{in}(u_f \odot h_{t-1}) \\
		& o_t = \sigma_{in}(u_o \odot  h_{t-1}) \\
		& \tilde{c_t} = \sigma(W_c x_t + U_c h_{t-1} + b_c) \\
		& c_t= f_t \odot c_{t-1} + i_t \odot \tilde{c_t}\\
		& h_t = o_t \odot \sigma(c_t) 
	\end{split}
\end{equation}

\subsection{LSTM5}
In the fifth model variant, we revise LSTM1 so that the matrices $U_i, U_f, U_o$ are replaced with corresponding vectors denoted by small letters. Then, as in LSTM4, we acquire (Hadamard) point-wise multiplication in the state variables. 
\begin{equation}
	\begin{split}
		& i_t = \sigma_{in}(u_i \odot h_{t-1} + b_i) \\
		& f_t = \sigma_{in}(u_f \odot  h_{t-1} + b_f) \\
		& o_t = \sigma_{in}(u_o \odot h_{t-1} + b_o) \\
		& \tilde{c_t} = \sigma(W_c x_t + U_c h_{t-1} + b_c) \\
		& c_t= f_t \odot c_{t-1} + i_t \odot \tilde{c_t}\\
		& h_t = o_t \odot \sigma(c_t) 
	\end{split}
\end{equation}
We note that for ease of tracking, odd-numbered variations contain biases while even-numbered variations do not. 

Table~\ref{vs} provides a summary of the number of parameters as well as the times per epoch during training corresponding to each of the 5 model variants.  We also add the parameter of the forward layer. These simulation and the training times are obtained by running the Keras Library \cite{keras_rowwise}.

\begin{table}
	\caption{variants specifications.}
	\centering
	\begin{tabular}{| c | c | c |} 
		\hline
		variants & \# of parameters & times(s) per epoch \\ 
		\hline
		LSTM & 52610 & 30 \\
		\hline
		LSTM1 & 44210 & 27 \\
		\hline
		LSTM2 & 43910 & 25 \\
		\hline
		LSTM3 & 14210 & 14 \\
		\hline
		LSTM4 & 14210 & 23 \\
		\hline
		LSTM5 & 14510 & 24 \\
		\hline
	\end{tabular}
	\label{vs}
\end{table}

\FloatBarrier
\section{Experiments and Discussion}
The goal of this paper is to provide a fair comparison among the five model variants and the standard LSTM model. We train and evaluate all models on the benchmark MNIST dataset using the images as row-wise sequence. MNIST images are $28 \times 28$. In the experiment, each model reads one row at a time from top to bottom to produce its output after seeing all $28$ rows. Table~\ref{tab:a} gives specification of network used. 
\begin{table}[ht]
	\caption{Network specifications.}
	\centering
	\begin{tabular}{| c | c |} 
		\hline
		Input dimension & $28 \times 28 $ \\ 
		\hline
		Number of hidden units & 100 \\
		\hline
		Non-linear function & tanh, sigmoid, tanh \\
		\hline
		Output dimension & 10 \\
		\hline
		Non-linear function & softmax \\
		\hline
		Number of epochs / Batch size & $100 / 32 $ \\
		\hline
		Optimizer /  Loss function & RMprop  /  categorical cross-entropy\\
		\hline
	\end{tabular}
	\label{tab:a}
\end{table}

Three different nonlinearities, i.e., $tanh, sigmoid,$ and $relu$,  have been employed of first (RNN) layer. For each case, we train three different cases with different values of $\eta$. Two of those for each case are depicted in the figures below while the Tables below summarize all three results. 

\subsection{The tanh activation}
The activation $tanh$ has been used as the nonlinearity of the first hidden layer. To improve performance of the model, we perform parameter tuning over different values of the learning parameter $\eta$. From experiments, see the samples in Fig.1 and Fig.2, as well as Table 1, there is a small amount of fluctuation in the testing accuracy; however, all variants converge to above $98\%$. The general trend among all three $\eta$ values is that LSTM1 and LSTM2 have the closest prediction to the standard LSTM. Then LSTM5 follows and finally LSTM4 and LSTM3. 
As it is shown, setting $\eta = 0.002$ results in test accuracy score of $98.60\%$ in LSTM3 (i.e., the fastest model with least number of parameters) which is close to the best test score of the standard LSTM, i.e., $99.09\%$. The best results obtained among all the epochs are shown in Table~\ref{tanh}. For each model, the best result over the 100 epochs training and using parameter tuning is shown in bold.

\begin{figure}[!htb]
	\vspace{-7mm}
	\setlength{\belowcaptionskip}{-20pt}
	\centering
	\includegraphics[trim={0 0 0 0},clip,scale=0.42]{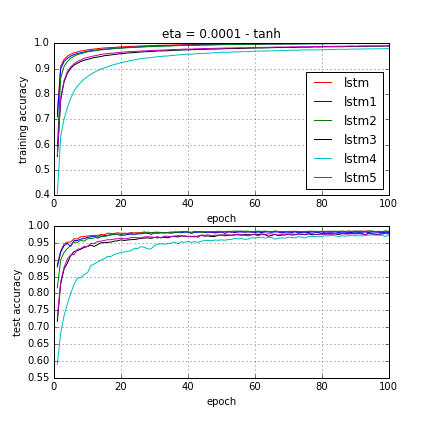}
	\caption{Training \& Test accuracy, $\sigma=tanh , \eta=1\mathrm{e}{-4}$}
	\label{fig:fig1}
\end{figure}

\begin{figure}[!htb]
	\centering
	\includegraphics[trim={0 0 0 0},clip,scale=0.42]{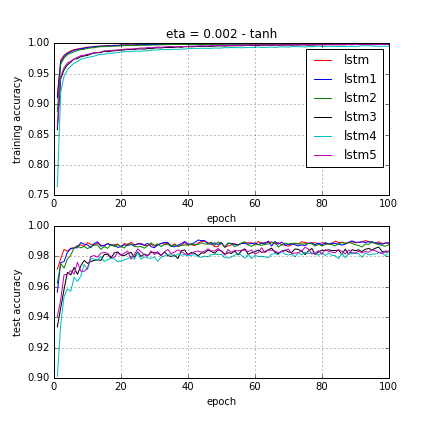}
	\caption{Training \& Test accuracy, $\sigma=tanh , \eta=2\mathrm{e}{-3}$}
	\label{fig:fig3}
\end{figure}

\begin{table}[!htb]
	\caption{Best results obtained Using tanh.}
	\centering
	\begin{tabular}{ |c|c|c|c|c| }
		\cline{3-5}
		\multicolumn{1}{c}{}
		& & $\eta = 1\mathrm{e}{-4}$ & $\eta =1\mathrm{e}{-3}$ & $\eta = 2\mathrm{e}{-3}$ \\
		\hline
		\multirow{2}{*}{LSTM} & train &  0.9995 &  1.0000 &  0.9994 \\ 
		& test &  0.9853 &  \textbf{0.9909} &  0.9903 \\ 
		\hline
		\multirow{2}{*}{LSTM1} & train &  0.9993 &  0.9999 &  0.9996 \\ 
		& test &  0.9828 &  0.9906 &  \textbf{0.9907} \\ 
		\hline
		\multirow{2}{*}{LSTM2} & train &  0.999 &  0.9997 &  0.9995 \\ 
		& test &  0.9849 &  0.9897 &  \textbf{0.9897} \\ 
		\hline
		\multirow{2}{*}{LSTM3} & train &  0.9889 &  0.9977 &  0.9983 \\ 
		& test &  0.9781 &  0.9827 &  \textbf{0.9860} \\ 
		\hline
		\multirow{2}{*}{LSTM4} & train &  0.9785 &  0.9975 &  0.9958 \\ 
		& test &  0.9734 &  \textbf{0.9853} &  0.9834 \\ 
		\hline
		\multirow{2}{*}{LSTM5} & train &  0.9898 &  0.9985 &  0.9983 \\ 
		& test &  0.9774 &  0.9835 &  \textbf{0.9859} \\ 
		\hline
	\end{tabular}
	\label{tanh}
\end{table}

\subsection{The (logistic) sigmoid activation}
Then, the sigmoid activation has been used as the nonlinearity of the first hidden layer. Again, we explored 3 different value of the learning parameter $\eta$. The same trend is observed using the sigmoid nonlinearity. In this case, one can clearly observe the training profile of each model. LSTM1, LSTM2, LSTM5, LSTM4 and LSTM3 have the closest prediction to the base LSTM respectively. Again larger $\eta$ results in better test accuracy and more fluctuation. It is observed that setting $\eta = 0.002$ results in test score of $98.34\%$ in LSTM3 which is close to the test score of base LSTM $98.86\%$. The best results obtained over the 100 epochs are summarized in Table~\ref{sigmoid}. \begin{figure}[ht]
	\vspace{-5mm}
	\centering
	\setlength{\belowcaptionskip}{-20pt}
	\includegraphics[trim={0 0 0 0},clip,scale=0.42]{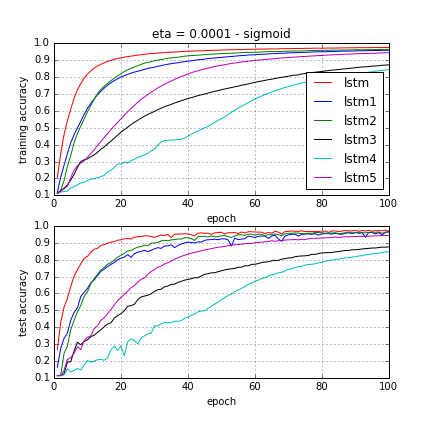}
	\caption{Training \& Test accuracy, $\sigma=sigmoid , \eta=1\mathrm{e}{-4}$}
	\label{fig:fig4}
\end{figure}

\begin{figure}[!htb]
	\vspace{-1mm}
	\centering
	\setlength{\belowcaptionskip}{-10pt}
	\includegraphics[trim={0 0 0 0},clip,scale=0.42]{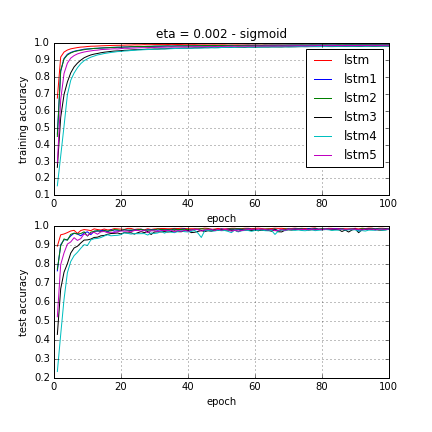}
	\caption{Training \& Test accuracy, $\sigma=sigmoid , \eta=2\mathrm{e}{-3}$}
	\label{fig:fig6}
\end{figure}

\begin{table}[!htb]
		\caption{Best results obtained sigmoid.}
		\centering
	\begin{tabular}{ |c|c|c|c|c| }
		\cline{3-5}
		\multicolumn{1}{c}{}
		& & $\eta = 1\mathrm{e}{-4}$ & $\eta =1\mathrm{e}{-3}$ & $\eta = 2\mathrm{e}{-3}$ \\
		\hline
		\multirow{2}{*}{LSTM} & train &  0.9751 &  0.9972 &  0.9978 \\ 
		& test &  0.9739 &  0.9880 &  \textbf{0.9886} \\ 
		\hline
		\multirow{2}{*}{LSTM1} & train &  0.9584 &  0.9901 &  0.9905 \\ 
		& test &  0.9635 &  \textbf{0.9863} &  0.9858 \\ 
		\hline
		\multirow{2}{*}{LSTM2} & train &  0.9636 &  0.9901 &  0.9907 \\ 
		& test &  0.9660 &  0.9856 &  \textbf{0.9858} \\ 
		\hline
		\multirow{2}{*}{LSTM3} & train &  0.8721 &  0.9787 &  0.9828 \\ 
		& test &  0.8757 &  0.9796 &  \textbf{0.9834} \\ 
		\hline
		\multirow{2}{*}{LSTM4} & train &  0.8439 &  0.9793 &  0.9839 \\ 
		& test &  0.8466 &  0.9781 &  \textbf{0.9822} \\ 
		\hline
		\multirow{2}{*}{LSTM5} & train &  0.9438 &  0.9849 &  0.9879 \\ 
		& test &  0.9431 &  0.9829 &  \textbf{0.9844} \\ 
		\hline
	
	\end{tabular}
	\label{sigmoid}
\end{table}

\subsection{The relu activation}
The $relu$ activation has been used as the nonlinearity of the first hidden layer. It is observed (in Fig. 6) that the performance of LSTM, LSTM1 and LSTM2 drop after a number of epochs; however, this is not the case for LSTM3, LSTM4 and LSTM5. These latter model are sustained for all three choices of $\eta$. Also LSTM3, the fastest model with least number of parameters, shows the best performance among all 5 variants! With the $relu$ as nonlinearity, the models fluctuate for larger $\eta$ which is not within the tolerance range of the model. Setting $\eta = 0.002$ results in test score of $99.00\%$ for LSTM3 which beat the best test score of the base LSTM, i.e., $98.43\%$. The best results obtained for all models are summarized in table~\ref{relu}. 

\begin{figure}[!htb]
	\vspace{-4mm}
	\centering
	\setlength{\belowcaptionskip}{-8pt}
	\includegraphics[trim={0 0 0 0},clip,scale=0.42]{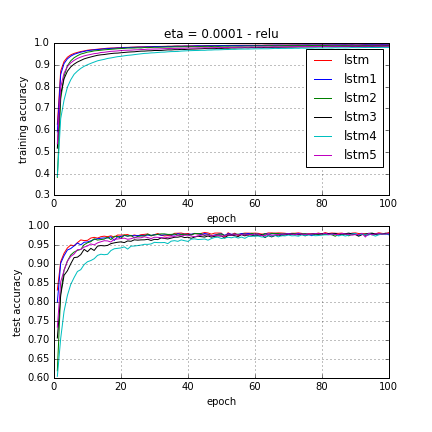}
	\caption{Training \& Test accuracy, $\sigma=relu , \eta=1\mathrm{e}{-4}$}
	\label{fig:fig7}
\end{figure}

\begin{figure}[!htb]
	\vspace{-4mm}
	\centering
	\setlength{\belowcaptionskip}{-5pt}
	\includegraphics[trim={0 0 0 0},clip,scale=0.42]{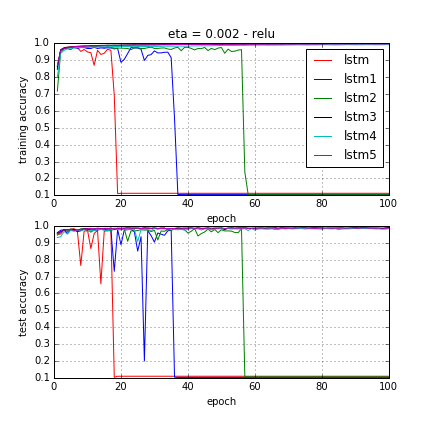}
	\caption{Training \& Test accuracy, $\sigma=relu , \eta=2\mathrm{e}{-3}$}
	\label{fig:fig9}
\end{figure}

\begin{table}[!htb]
	\caption{Best results obtained by relu model.}
	\centering
	\begin{tabular}{ |c|c|c|c|c| }
		\cline{3-5}
		\multicolumn{1}{c}{}
		& & $\eta = 1\mathrm{e}{-4}$ & $\eta =1\mathrm{e}{-3}$ & $\eta = 2\mathrm{e}{-3}$ \\
		\hline
		\multirow{2}{*}{LSTM} & train &  0.9932 &  0.9829 &  0.9787 \\ 
		& test &  0.9824 &  \textbf{0.9843} &  0.9833 \\ 
		\hline
		\multirow{2}{*}{LSTM1} & train &  0.9926 &  0.9824 &  0.9758 \\ 
		& test &  0.9803 &  \textbf{0.9832} &  0.9806 \\ 
		\hline
		\multirow{2}{*}{LSTM2} & train &  0.9896 &  0.9795 &  0.98 \\ 
		& test &  0.9802 &  0.9805 &  \textbf{0.9836} \\ 
		\hline
		\multirow{2}{*}{LSTM3} & train &  0.9865 &  0.9967 &  0.9968 \\ 
		& test &  0.9808 &  0.9882 &  \textbf{0.9900} \\ 
		\hline
		\multirow{2}{*}{LSTM4} & train &  0.9808 &  0.9916 &  0.9918 \\ 
		& test &  0.9796 &  \textbf{0.9857} &  0.9847 \\ 
		\hline
		\multirow{2}{*}{LSTM5} & train &  0.987 &  0.9962 &  0.9964 \\ 
		& test &  0.9807 &  0.9885 &  \textbf{0.9892} \\ 
		\hline
	\end{tabular}
	\label{relu}
\end{table}

\section{Conclusion}
Five variants of the base LSTM model has been presented and evaluated. These models have been examined and evaluated on the benchmark classical MNIST dataset using different nonlinearity and different learning rates $\eta$. In the first model variant, the input  and their weights have been removed uniformaly from the three gates. In the second model variant, the input weight and the bias have been removed from all gates. In the third model, the gates only retain their biases. The fourth model variant is similar to the second variant, and fifth variant is similar to first variant, except that weights become vectors to execute point-wise multiplication. It has been found that new model variants are comparable to the base LSTM model. Thus, these varaint models may be suitably chosen in applications in order to benefit from speed and/or computational cost.
\section*{Acknowledgment}
This work was supported in part by the National Science Foundation under grant No. ECCS-1549517.
\small{
	\bibliographystyle{ieee}
	\bibliography{egbib}
}

\end{document}